\pdfoutput=1

\documentclass[11pt]{article}

\usepackage[final]{acl}

\usepackage{times}
\usepackage{latexsym}
\usepackage{enumitem}
\usepackage[T1]{fontenc}

\usepackage[utf8]{inputenc}

\usepackage{microtype}

\usepackage{inconsolata}

\usepackage{graphicx}

\usepackage{amsmath}
\usepackage{amssymb}
\usepackage{multirow}
\usepackage{hyperref}
\usepackage{kotex}
\usepackage[normalem]{ulem}
\useunder{\uline}{\ul}{}
\usepackage{color, colortbl}
\usepackage{booktabs}
\usepackage{graphicx}
\usepackage{adjustbox}
\usepackage{subcaption}
\usepackage{authblk}
\usepackage{url}

\definecolor{chembrayblue}{rgb}{0.6196, 0.7059 0.8275}
\definecolor{lightblue}{rgb}{0.886, 0.929, 0.996}
\definecolor{gray}{rgb}{0.937, 0.937, 0.937}

\title{Bridging the Gap Between Molecule and Textual Descriptions via\\Substructure-aware Alignment}

\author{
\textbf{Hyuntae Park\textsuperscript{1}}\thanks{These authors contributed equally to this work.} \quad
\textbf{Yeachan Kim\textsuperscript{2*}} \quad
\textbf{SangKeun Lee\textsuperscript{1,3}} \\
\textsuperscript{1}Department of Artificial Intelligence, Korea University, Seoul, Republic of Korea \\
\textsuperscript{2}Division of Language \& AI, Hankuk University of Foreign Studies, Seoul, Republic of Korea \\
\textsuperscript{3}Department of Computer Science and Engineering, Korea University, Seoul, Republic of Korea \\
\texttt{pht0639@korea.ac.kr, yeachan@hufs.ac.kr, yalphy@korea.ac.kr}
}

\begin{document}
\maketitle
\begin{abstract}
Molecule and text representation learning has gained increasing interest due to its potential for enhancing the understanding of chemical information. However, existing models often struggle to capture subtle differences between molecules and their descriptions, as they lack the ability to learn fine-grained alignments between molecular substructures and chemical phrases. To address this limitation, we introduce MolBridge, a novel molecule–text learning framework based on substructure-aware alignments. Specifically, we augment the original molecule–description pairs with additional alignment signals derived from molecular substructures and chemical phrases. To effectively learn from these enriched alignments, MolBridge employs substructure-aware contrastive learning, coupled with a self-refinement mechanism that filters out noisy alignment signals. Experimental results show that MolBridge effectively captures fine-grained correspondences and outperforms state-of-the-art baselines on a wide range of molecular benchmarks, underscoring the importance of substructure-aware alignment in molecule-text learning.\footnote{Our code and data are available at \url{https://github.com/Park-ing-lot/MolBridge}}

\end{abstract}

\section{Introduction}

Recent advances in natural language processing (NLP) have transformed various scientific fields, with chemistry emerging as a prominent domain. Transformer-based models have demonstrated remarkable success in molecular tasks, such as drug discovery \citep{drews2000drug} and molecular property prediction \citep{wu2018moleculenet}, offering scalable alternatives to traditional wet-lab experiments \citep{scibert, smilesbert}. Among these advancements, Molecule-Text Models (MTMs) have been developed to bridge the gap between molecular structures and natural language, providing a symbolic interface for understanding complex chemical information \citep{molt5, moleculeSTM, molca}.

Despite their potential, MTMs face a fundamental challenge: the severe \textbf{sparsity} of molecule-text alignment. Given the vast diversity of chemical structures, annotated datasets that explicitly pair molecules with their corresponding textual descriptions are extremely limited. This scarcity restricts the model’s ability to generalize across chemical space, leading to biased representations that perform poorly on unseen compounds \cite{haghighatlari2020learning}. 
More critically, it prevents MTMs from learning fine-grained correspondences between specific fragments (i.e., molecular substructures and corresponding chemical phrases), making them struggle to capture subtle differences between similar compounds, such as D-glutamate and L-glutamate \citep{zhang2025atomas}. These subtle differences reflect substructural variations, which are important because they often determine the core functionalities and chemical properties of the entire molecule \citep{wu2023chemistry}.

Although some studies \citep{DBLP:conf/bibm/MinLZSSHB24, zhang2025atomas} have attempted to address this issue by introducing local alignments on the given pairs, these methods still suffer from several limitations. 
First, they rely heavily on \textbf{indirect alignment}, where local relations are inferred through feature similarity due to the lack of explicit fragment-level annotations. This absence of direct supervision can lead to incorrect or incomplete mappings, making it difficult for models to learn accurate local relationships between fragments.
Second, they often suffer from \textbf{over-fragmented alignment}, where models attempt to align molecular tokens (e.g., SMILES characters like `=', `[]', `()') indiscriminately. Such token-level alignment introduces noise, causing the model to learn semantically meaningless fragments rather than chemically meaningful substructures. These limitations comprehensively hinder the ability of existing MTMs to achieve accurate fine-grained alignment and robust generalization.


In response, we propose MolBridge, a novel multimodal framework designed to learn fine-grained alignment between molecules and text through substructure-aware alignments.
MolBridge begins by explicitly extracting substructures from molecules and chemical phrases from their corresponding descriptions. These fragments are then cross-linked to their semantically or chemically relevant counterparts: chemical phrases are associated with entire molecules, while substructures are connected to descriptions. To effectively learn from these enriched alignments, we introduce substructure-aware contrastive learning, which jointly considers both fragment-level and holistic molecule-text relations. 
This strategy encourages the model to capture meaningful substructural semantics while preserving consistency between molecules and their descriptions. 

Building on the substructure-aware representations learned by MolBridge, we also introduce MolBridge-Gen, a generative variant of the framework that explicitly leverages local alignment signals derived from substructure–chemical phrase pairs identified by MolBridge. This extension enables the framework to generalize beyond discriminative tasks and effectively support generative scenarios, such as molecule captioning and generation, where fine-grained semantic understanding is essential \cite{DBLP:conf/nips/XiaZZ0GH0ZLL23}.

We conduct comprehensive evaluations across core molecular tasks, including molecular property prediction \citep{wu2018moleculenet}, molecule–text retrieval \citep{kvplm, molca}, and generation tasks \citep{molt5}, to thoroughly assess the effectiveness of MolBridge.
Experimental results show that MolBridge consistently outperforms existing MTMs, achieving superior fine-grained alignment accuracy, higher retrieval performance, and enhanced generation quality. The contributions of this paper include the followings:
\begin{itemize} 
\itemsep0em
    \item We propose MolBridge, a novel framework for fine-grained molecule-text alignment, directly addressing the sparsity of alignment datasets through substructure-aware alignments.
    
    \item We introduce a substructure-aware contrastive learning, allowing the model to effectively capture fine-grained relations between molecules and text descriptions.
    
    \item We demonstrate that MolBridge consistently outperforms existing methods on diverse tasks, highlighting the significance of substructure-aware augmentations.
\end{itemize}

\begin{figure*}[t]
\centering
\subfloat[Training procedure of MolBridge. We first extract substructures and chemical phrases from SMILES and caption, respectively. We then augment our dataset by constructing additional positive pairs based on their substructural relationships and perform substructure-aware contrastive learning.]{%
  \includegraphics[width=\linewidth]{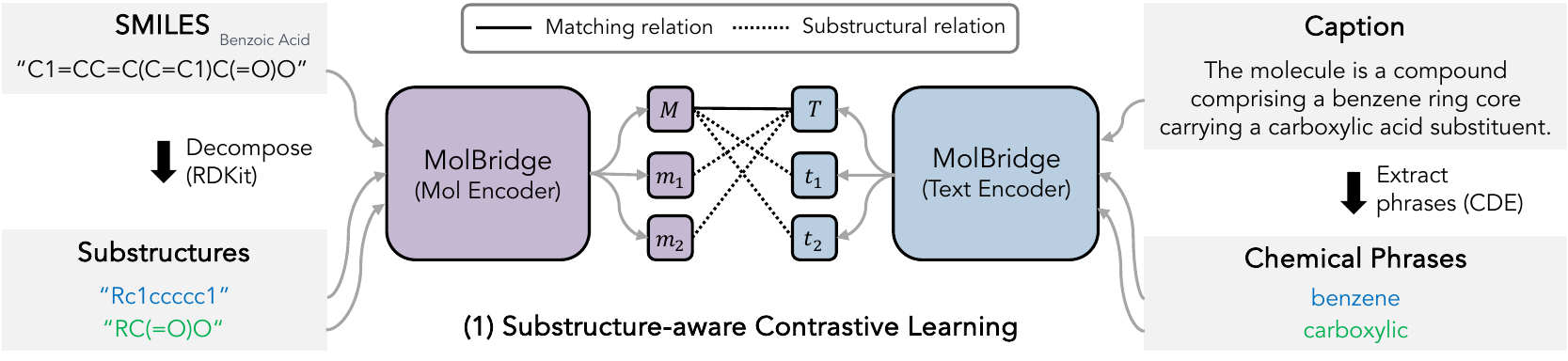}%
  \label{molbridge}
}

\subfloat[Training procedure of MolBridge-Gen. After identifying local relations using MolBridge, we train the generative molecule-text models with them.]{%
  \includegraphics[width=\linewidth]{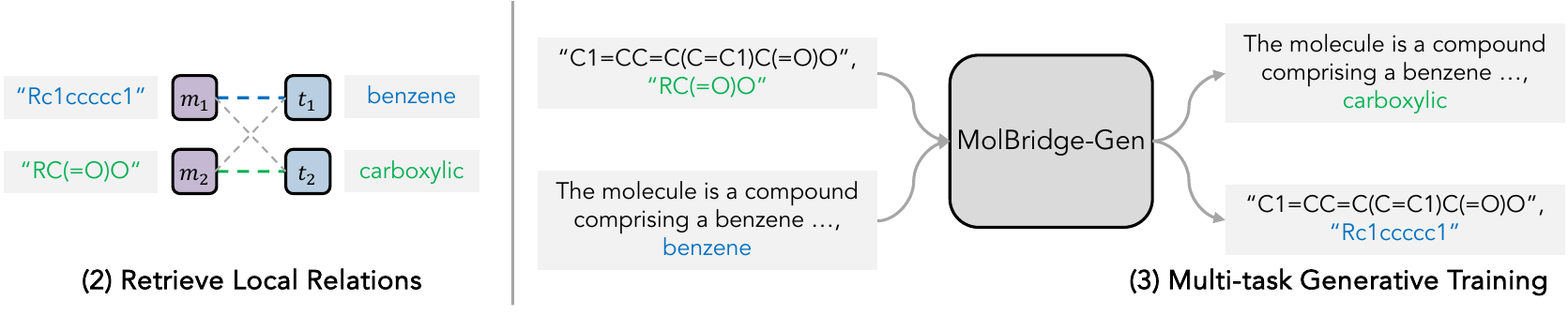}%
  \label{molbridge-gen}
}
\caption{Illustration of our framework to learn fine-grained alignments.}
\label{fig:model}
\end{figure*}

\section{Related Works}
\subsection{Molecule-Text Multimodal Modeling}
Various models have emerged for learning molecule-text representations. Early on, \citet{kvplm} attempted to leverage english textual knowledge to enhance molecule representation learning. \citet{moleculeSTM} proposed MoleculeSTM, applying larger datasets and contrastive learning to improve alignment between molecules and text. To facilitate the translation between molecules and natural language, \citet{molt5} proposed molecule captioning tasks along with a powerful baseline model, MolT5. Inspired by \citet{DBLP:journals/jmlr/RaffelSRLNMZLL20}, they utilized corrupted spans replacement objective for pre-training to improve molecular understanding. \citet{DBLP:conf/acl/LiuZXW0QZL23} employed generative pre-training to reflect the importance of descriptions, while \citet{molca} introduced MolCA, a molecular text model that leverages a 2D graph for enhanced molecular understanding. Recently, large language model-based instruction-tuned generative models for molecule understanding \citet{DBLP:conf/iclr/FangL0LH0FC24-molinstruction, DBLP:conf/acl/PeiWGLFZ00024-biot5plus, DBLP:conf/coling/CaoLLYL25-instructmol} have been proposed to enhance generalizability through multitask learning. However, these works only consider the global representations of molecules and text, overlooking finer-grained modal interactions. Unlike these approaches, we encourage the model to effectively learn compositional structures in both molecules and language.

\subsection{Fine-grained Molecule-Text Representation Learning }
Several studies have attempted to perform fine-grained molecule-text alignment. \citet{DBLP:conf/iclr/Yu0NFLZL24-molblend, DBLP:journals/corr/abs-2310-14216-unimap} have focused on fine-grained alignment of molecular modalities and tailoring for prediction tasks. Unlike these paradigms, Atomas \citep{zhang2025atomas} tackles the problem of cross-modal learning between molecular structures and textual descriptions. A key challenge is the scarcity of fine-grained expert annotations and the ambiguity in defining positive and negative pairs, since one textual property may relate to multiple substructures. To overcome this, the authors use clustering techniques to learn multi-scale representations and encourage consistency across different levels of granularity. Similarly, \citet{DBLP:conf/bibm/MinLZSSHB24} utilized optimal transport to conduct fragments at the atom, motif, and global levels within the embedding space. 
Unlike these methods for implicit local alignment, \citet{DBLP:journals/corr/abs-2411-14721} proposed an explicit local alignment approach; however, it heavily relies on costly large language models. 
In this work, we address the challenge of explicit fine-grained alignment by learning fragment-level representations enriched with substructural cues, such as identifying whether a chemical phrase is part of a molecule or a substructure is mentioned in the description, and discovering fine-grained correspondences between these fragments.

\section{Methodology}

In this section, we introduce MolBridge, a novel framework designed to learn fine-grained alignments between molecules and its descriptions. MolBridge begins with a sparse molecule–text alignment dataset, from which it extracts chemical substructures and phrases to form explicit alignment pairs (\S\ref{method:data}). These augmented pairs then guide structure-aware contrastive learning, enhanced by a self-refinement procedure (\S\ref{method:pretrain}). Finally, we extend the learned representations to various molecular tasks by either directly adopting MolBridge or combining it with generative models (\S\ref{method:expansion}). The overall procedure is illustrated in Figure \ref{fig:model}.

\subsection{Substructure Alignment Augmentation}
\label{method:data}

Given the vast diversity of chemical space, existing alignment datasets remain severely limited, constraining a model’s ability to generalize across novel molecules \cite{haghighatlari2020learning}. To overcome this challenge, we exploit the inherent substructural relations between molecules and their textual descriptions. Let $S = (M, T)$ denote a molecule–text pair, where $M$ is represented by its SMILES \cite{weininger1988smiles} string and $T$ is the corresponding caption, we aim to enrich the following alignments:
\paragraph{Substructures ($m$) to Description ($T$)}

The chemical properties of molecules are largely determined by their constituent substructures (e.g., functional groups, ring systems), indicating the potential alignment between the substructure and original molecule's description. To capture these relations, we decompose each original molecule into a set of substructures using established fragmentation methods, including BRICS \cite{degen2008art}, RECAP \cite{lewell1998recap} fragmentation methods\footnote{We compare various decomposition strategies in Table \ref{tab:rdkit} of the Appendix.}. We then link each substructure $m$ to the original description, yielding a set of substructure–text alignments $(m,T) \in S_{m}$. These enriched pairs are subsequently integrated into the original alignment set $S$.

\paragraph{Chemical Phrases ($t$) to Molecules ($M$)}

Additionally, we align chemical phrases extracted from each description with the corresponding molecule. This step is motivated by the observation that molecular captions often contain non-informative tokens—such as articles, pronouns, and filler words—that are unrelated to chemical properties and may introduce noise into the alignment process \cite{DBLP:conf/icml/RadfordKHRGASAM21,messina2021fine}. To this end, we extract chemical phrases $t$ from the original caption $T$ using two approaches: (i) ChemDataExtractor \cite{DBLP:journals/jcisd/MavracicCIEC21_cem2}, a chemistry-specific phrase extractor, and (ii) a large language model–based method\footnote{We compare these two methods in Section \ref{extractor} of the Appendix and, for our main experiments, adopt ChemDataExtractor due to its superior cost-effectiveness}. As with substructures, each phrase is then linked back to the original molecule, yielding molecule–phrase alignments $(M,t) \in S_{t}$. Notably, these chemical phrases can also provide contextual cues for understanding molecular substructures, as many phrases directly reference functional groups (e.g., hydroxyl aromatic), thereby enhancing substructure-aware alignment. These enriched pairs are finally added to the alignment set $S$.

\subsection{Substructure-aware Contrastive Learning with Self-Refinement}
\label{method:pretrain}

Based on the augmented datasets, we train MTMs to learn fine-grained alignments. Considering that the augmented alignments are structured in one-to-many relations, we introduce a substructure-aware contrastive learning that explicitly aligns (i) Molecules with both their original descriptions and extracted chemical phrases (ii) Descriptions with both their original molecules and associated chemical substructures. However, the augmented alignment set potentially involves the incorrect alignment. To address this, the substructure-aware training is built on the self-refinement process.

\paragraph{Substructure-aware Contrastive Learning} Given a mini-batch of molecule-side inputs $x_m^i$ and text-side inputs $x_t^j$, we encode them with modality-specific encoders $f_m(\cdot)$ and $f_t(\cdot)$. We define the pairwise similarity:
\begin{equation}
\sigma_{i,j} = \exp{\left(\frac{1}{\tau}\cdot\frac{f_m(x_m^i)\cdot f_t(x_t^j)}{\|f_m(x_m^i)\|\|f_t(x_t^j)\|}\right)},
\end{equation}
where $\tau$ is a learnable temperature parameter, and we use the hidden state of the first token to compute similarity.

For each anchor $i$ in standard contrastive learning, let $\mathcal{P}(i)$ be its set of positive matches (i.e., aligned) drawn from our augmented alignments, and let $\mathcal{U}(i)$ be the set that includes contrastive examples in batches, excluding substructure-phrase pairs to avoid false negatives from semantically related examples. 
We then optimize the in-batch contrastive loss:
\begin{equation}
\mathcal{L}_{\text{mol2txt}} = -\frac{1}{N}\sum_{i=1}^{N}\frac{1}{|\mathcal{P}(i)|}\log\frac{\sum_{j\in\mathcal{P}(i)}\sigma_{i,j}}{\sum_{k\in\mathcal{U}(i)}\sigma_{i,k}},\nonumber
\end{equation}

\begin{equation}\label{eq:contrastive}
\mathcal{L}_{\text{txt2mol}} = -\frac{1}{N}\sum_{j=1}^{N}\frac{1}{|\mathcal{P}(j)|}\log\frac{\sum_{i\in\mathcal{P}(j)}\sigma_{i,j}}{\sum_{k\in\mathcal{U}(j)}\sigma_{k,j}},
\end{equation}
where $\mathcal{L}_{\text{mol2txt}}$ denotes the contrastive loss when molecules serve as anchors and its associated text units (descriptions or phrases) as positives, while $\mathcal{L}_{\text{txt2mol}}$ denotes the converse loss, with text units as anchors and molecular structures as positives. The total loss for substructure-aware contrastive learning is as follows:
\begin{equation}\label{eq:total_loss}
    \mathcal{L} =  \mathcal{L}_{\text{txt2mol}} + \mathcal{L}_{\text{mol2txt}}
\end{equation}

\paragraph{Self-Refinement} 
Although we consider potential relationships between molecules and their descriptions, some incorrect associations may still arise. To detect and remove these low-quality pairs, we embed our contrastive training within an iterative self-refinement loop.

To obtain signals for erroneous relations, we introduce a relation classification loss to the total loss (Eq.~\eqref{eq:total_loss}):
\begin{equation} 
\mathcal{L}_{cl} = -\frac{1}{N} \sum_{i=1}^{N} \sum_{c=1}^{3} y_{i,c} \log p_{i,c}(f_m(x_m^i)\oplus f_t(x_t^i)), 
\end{equation}
where $y_{i,c}$ is the ground-truth indicator for class $c\in\{S, S_m, S_t\}$, and $p_{i,c}$ is the model’s predicted probability that pair ($x_m^i, x_t^i$) belongs to class $c$. Inspired by the observation that \textit{models learn clean samples before noisy ones} \cite{arazo2019unsupervised}, we discard any pairs that are misclassified in all of a set of predefined epochs. This filtering ensures that subsequent training focuses on higher-quality alignment signals.

\begin{table*}[ht!]
\footnotesize
\centering
\begin{tabular}{@{}l|c|cccc|cccc@{}}
\toprule
\multirow{2}{*}{Methods} & \multirow{2}{*}{\# Params} & \multicolumn{4}{c|}{Text to Molecule} & \multicolumn{4}{c}{Molecule to Text} \\
 & & R@1 & R@5 & R@10 & MRR & R@1 & R@5 & R@10 & MRR \\ \midrule
\rowcolor{gray}
\multicolumn{10}{l}{1D SMILES + 2D Graph} \\
MoMu \citep{momu} & 111M & 4.90 & 14.48 & 20.69 & 10.33 & 5.08 & 12.82 & 18.93 & 9.89 \\
MolFM \citep{molfm} & 138M & 16.14 & 30.67 & 39.54 & 23.63 & 13.90 & 28.69 & 36.21 & 21.42 \\
MolFM (fine-tuned) \citep{molfm} & 138M & 29.39 & 50.26 & 58.49 & 39.34 & 29.76 & 50.53 & 58.63 & 39.56 \\
MolCA \citep{molca} & 111M & 35.09 & 62.14 & 69.77 & 47.33 & 37.95 & 66.81 & 74.48 & 50.80 \\ \midrule
\rowcolor{gray}
\multicolumn{10}{l}{1D SMILES} \\
MoleculeSTM \citep{moleculeSTM} & 120M & 35.80 & - & - & - & 39.50 & - & - & - \\
Atomas-base \citep{zhang2025atomas} & 271M & 39.08 & 59.72 & 66.56 & 47.33 & 37.88 & 59.22 & 65.56 & 47.81 \\
Atomas-large \citep{zhang2025atomas} & 825M & 49.08 & 68.32 & 73.16 & 57.79 & 46.22 & 66.02 & 72.32 & 55.52 \\
\rowcolor{lightblue}
MolBridge w/o augmentation & 155M & 23.89 & 48.91 & 57.53 & 35.30 & 27.30 & 51.86 & 60.34 & 38.47 \\
\rowcolor{lightblue}
MolBridge & 155M & \textbf{50.45} & \textbf{70.83} & \textbf{76.11} & \textbf{59.63} & \textbf{52.76} & \textbf{73.54} & \textbf{78.55} & \textbf{62.25} \\ \bottomrule
\end{tabular}
\caption{Zero-shot molecule-text retrieval performance on PCDes test set (scaffold split). w/o augmentation refers to the model trained without substructure alignment augmentation. Baseline results are from \citet{zhang2025atomas}.}
\label{tab:ret1}
\end{table*}

\begin{table}[ht!]
\centering
\resizebox{\columnwidth}{!}{%
\begin{tabular}{@{}l|cc|cc@{}}
\toprule
\multirow{2}{*}{Methods} & \multicolumn{2}{c|}{T2M} & \multicolumn{2}{c}{M2T} \\
 & R@1 & R@20 & R@1 & R@20 \\ \midrule
\rowcolor{gray}
\multicolumn{5}{l}{1D SMILES + 2D Graph} \\
MoMu-S \citep{momu} & 40.8 & 86.1 & 40.9 & 86.2 \\
MoMu-K \citep{momu} & 41.6 & 87.8 & 41.8 & 87.5 \\
MoleculeSTM \citep{moleculeSTM} & 44.3 & 90.3 & 45.8 & 88.4 \\
MolCA \citep{molca} & 66.0 & 93.5 & 66.6 & 94.6 \\ \midrule
\rowcolor{gray}
\multicolumn{5}{l}{1D SMILES} \\
SciBERT \citep{scibert} & 37.5 & 85.2 & 39.7 & 85.8 \\
KV-PLM \citep{kvplm}& 37.7 & 85.5 & 38.8 & 86.0 \\
\rowcolor{lightblue}
MolBridge & \textbf{70.9} & \textbf{95.6} & \textbf{75.0} & \textbf{97.4} \\ \bottomrule
\end{tabular}
}
\caption{Zero-shot molecule-text retrieval performance on Pubchem324k test set. Baseline results are from \citet{molca}.}
\label{tab:ret2}
\end{table}
 
\subsection{MolBridge with Generative Models}\label{method:expansion}

For molecular generative tasks (e.g., molecule captioning, molecule generation), we need to train generative models with the translation objective. While the previous augmented datasets provide valuable insights into the fine-grained alignment between molecules and descriptions, they are not directly applicable to this translation task, which demands one-to-one mappings at the same semantic level (i.e., molecule-to-description $(M,T)$ or substructure-to-phrases $(m,t)$). 

\paragraph{Substructure–phrase Relations} Accurately identifying these one-to-one relations is challenging due to the absence of explicit supervision linking substructures and phrases. To address this, we leverage the pre-trained MolBridge, which has been trained on the previously augmented datasets. This model inherently captures fine-grained associations between substructures and phrases through the substructure-aware alignment.

To obtain these relations, we begin by extracting substructures and phrases from the original training dataset. Each substructure is then paired with candidate phrases, and the relevance of each pair is evaluated using the MolBridge score—defined as the cosine similarity between substructure and phrase embeddings. Only substructure–phrase pairs with scores exceeding a predefined threshold $\tau$ are retained, ensuring high-quality alignment. If no valid pair is found for a given molecule, the original pair is excluded from training, as it lacks sufficient alignment signals.

\paragraph{Training MolBridge-Gen}
The resulting substructure–phrase pairs are used to train MolBridge-Gen, a generative model optimized using a conditional generation loss in a multi-task setting, following \citet{DBLP:conf/icml/Christofidellis23_t5+chem}. For example, in the molecule captioning task, MolBridge-Gen is trained to simultaneously generate full captions from the complete molecular representation and chemical phrases from the substructures. This dual-generation strategy ensures that the model learns both the original context of the molecule and the fine-grained details of its substructures. Detailed prompt templates used for pre-training are provided in Table \ref{tab:prompts} in the Appendix.

\begin{table*}[ht!]
\footnotesize
\centering
\resizebox{\linewidth}{!}{%
\begin{tabular}{@{}lccccccccc@{}}
\toprule
Method & BBBP & Tox21 & ToxCast & ClinTox & MUV & HIV & BACE & SIDER & Avg. \\ \midrule
MoleculeSTM \citep{moleculeSTM} & 70.6 & 75.7 & 65.2 & 86.6 & 65.7 & 77.0 & 82.0 & 63.7 & 73.3 \\
MolFM \citep{molfm} & 72.9 & 77.2 & 64.4 & 79.7 & 76.0 & 78.8 & 83.9 & 64.2 & 74.6 \\
MoMu \citep{momu} & 70.5 & 75.6 & 63.4 & 79.9 & 70.6 & 75.9 & 76.7 & 60.5 & 71.6 \\
MolCA-SMILES \citep{molca} & 70.8 & 76.0 & 56.2 & 89.0 & - & - & 79.3 & 61.1 & - \\
Atomas \citep{zhang2025atomas} & 73.7 & 77.9 & 66.9 & 93.2 & 76.3 & \textbf{80.6} & 83.1 & 64.4 & 77.0 \\ 
\rowcolor{lightblue}
\rowcolor{lightblue}
MolBridge & \textbf{77.6} & \textbf{84.7} & \textbf{70.3} & \textbf{94.8} & \textbf{76.8} & 77.8 & \textbf{84.5} & \textbf{66.9} & \textbf{79.2} \\ \bottomrule
\end{tabular}%
}
\caption{Results for molecular property prediction tasks (ROC-AUC) on MoleculeNet benchmark. \textbf{Bold} indicates the best results.}
\label{tab:property}
\end{table*}

\section{Experiments}
In this section, we verify the efficacy of MolBridge through extensive experiments and analyses aimed at answering the following questions:

\begin{itemize}[leftmargin=*]
\itemsep0em
\item[$\circ$] Can MolBridge capture fine-grained alignments more effectively than existing MTMs? (\S\ref{retrieval})

\item[$\circ$] Can MolBridge transfer its learned representations to uncover diverse structure–property relationships in downstream tasks? (\S\ref{property})

\item[$\circ$] Can the relations identified by MolBridge yield interpretable alignments that support effective translation between molecules and text?
    (\S\ref{captioning})
    
\end{itemize}

\subsection{Experimental Settings}
\paragraph{Dataset.} 
For training MolBridge, we collect the descriptions for 431,877 molecules following previous works \citep{moleculeSTM, molca}, removing any data overlapping with downstream task datasets to prevent data leakage. The augmented dataset contains approximately 2M pairs.
We train MolBridge-Gen with 32,455 pairs of data that are estimated to contain local relations as described in Section \ref{method:expansion}.
When decomposing molecules into substructures, we set a maximum number of atoms to 100 due to its high computational complexity. We extract chemical phrases from all captions in the dataset.
For evaluation, we perform zero-shot molecule-text retrieval tasks on the PubChem324k \citep{molca} and PCdes \citep{kvplm} datasets, molecule captioning tasks on the ChEBI-20 \citep{DBLP:conf/emnlp/EdwardsZJ21} dataset, and molecule property prediction tasks using the MoleculeNet benchmark \citep{wu2018moleculenet}. Details of the implementation are provided in Appendix \ref{implementation}.

\subsection{Zero-shot Molecule-Text Retrieval}
\label{retrieval}
\paragraph{Settings.} 
We report zero-shot retrieval performance using Recall at 1/5/10/20, which measures the proportion of relevant results found within the top 1, 5, 10, or 20 positions, a performance metric for information retrieval systems \citep{DBLP:books/daglib/0021593}. We also report the Mean Reversed Rank (MRR) \citep{DBLP:conf/trec/Voorhees99}, which measures how effectively a retrieval model ranks relevant items by averaging the inverse rank positions of the first correct result across multiple queries.

\paragraph{Results.} 
We evaluated retrieval performance on three datasets, as summarized in Tables~\ref{tab:ret1} and~\ref{tab:ret2}. On the PCDes scaffold test set, MolBridge (155M) achieves substantial performance gains compared to both Atomas-base (271M) and Atomas-large (825M), despite having significantly fewer parameters. Specifically, MolBridge shows average improvements of 11.1\%p and 14.2\%p over Atomas-base, and 2.2\%p and 6.8\%p over Atomas-large in text-to-molecule and molecule-to-text retrieval, respectively\footnote{Results on the original PCDes split are shown in Table~\ref{tab:ret3} in the Appendix.}. This result demonstrates (i) the efficiency and effectiveness of MolBridge in capturing fine-grained molecule–text alignments with a more compact architecture (ii) prior methods relying solely on implicit alignment signals may learn incorrect fragment correspondences.

Further, removing our substructural alignment augmentation leads to a noticeable drop in performance, validating its critical role in guiding fragment-level representation learning.
Table~\ref{tab:ret2} reports the performance on the PubChem324k test set, where MolBridge again outperforms all baselines, including those utilizing 2D molecular graphs, demonstrating its effectiveness in accurately linking molecular structures with natural language descriptions.

\begin{table*}[ht!]
\centering
\resizebox{\textwidth}{!}{%
\begin{tabular}{@{}lccccccc@{}}
\toprule
\multicolumn{1}{l|}{Method} & \multicolumn{1}{c|}{\# Params} & BLEU-2$\uparrow$ & BLUE-4$\uparrow$ & ROUGE-1$\uparrow$ & ROUGE-2$\uparrow$ & ROUGE-L$\uparrow$ & METEOR$\uparrow$ \\ \midrule
\multicolumn{1}{l|}{MolT5-large \citep{molt5}} & \multicolumn{1}{c|}{783M} & 0.594 & 0.508 & 0.654 & 0.510 & 0.594 & 0.614 \\
\multicolumn{1}{l|}{Text+Chem T5 \citep{DBLP:conf/icml/Christofidellis23_t5+chem}} & \multicolumn{1}{c|}{220M} & 0.625 & 0.542 & 0.682 & 0.543 & 0.622 & 0.648 \\
\multicolumn{1}{l|}{MolReGPT (GPT-4) \citep{molregpt}} & \multicolumn{1}{c|}{-} & 0.607 & 0.525 & 0.634 & 0.476 & 0.562 & 0.610 \\
\multicolumn{1}{l|}{Atomas-base \citep{zhang2025atomas}} & \multicolumn{1}{c|}{271M} & \uline{0.632} & \uline{0.545} & 0.685 & 0.545 & 0.626 & - \\
\multicolumn{1}{l|}{MolReFlect \citep{DBLP:journals/corr/abs-2411-14721}} & \multicolumn{1}{c|}{7B} & 0.617 & 0.539 & 0.657 & 0.510 & 0.593 & 0.623 \\
\rowcolor{lightblue}
\multicolumn{1}{l|}{MolBridge-Gen-small} & \multicolumn{1}{c|}{82M} & 0.625 & 0.542 & \uline{0.686} & \uline{0.549} & \uline{0.629} & \uline{0.649} \\ 
\rowcolor{lightblue}
\multicolumn{1}{l|}{MolBridge-Gen-base} & \multicolumn{1}{c|}{248M} & \textbf{0.674} & \textbf{0.605} & \textbf{0.724} & \textbf{0.609} & \textbf{0.676} & \textbf{0.693} \\ 
\bottomrule
\end{tabular}%
}
\caption{Results of molecule captioning task on CheBI-20 test set. \textbf{Bold} and \uline{underlined} indicate the best and second-best results, respectively. Full comparison is in Table \ref{tab:molcap_appendix} in the Appendix.}
\label{tab:molcap}
\end{table*}

\begin{table*}[ht!]
\centering
\resizebox{\textwidth}{!}{%
\begin{tabular}{@{}lccccccc@{}}
\toprule
Method & BLEU$\uparrow$ & EM$\uparrow$ & Levenshtein$\downarrow$ & MACCS FTS$\uparrow$ & RDK FTS$\uparrow$ & Morgan FTS$\uparrow$ & Validity$\uparrow$ \\ \midrule
MolT5-large \citep{molt5} & 0.854 & 0.318 & 16.32 & 0.889 & 0.813 & 0.750 & 0.958 \\
Text+Chem T5 \citep{DBLP:conf/icml/Christofidellis23_t5+chem} & 0.853 & 0.322 & 16.87 & 0.901 & 0.816 & 0.757 & 0.943 \\
MolReGPT (GPT-4) \citep{molregpt} & 0.857 & 0.280 & 17.14 & 0.903 & 0.805 & 0.739 & 0.899 \\
Atomas-large \citep{zhang2025atomas} & \textbf{0.874} & \textbf{0.387} & \textbf{12.70} & 0.914 & \uline{0.841} & \uline{0.788} & \textbf{0.980} \\
MolReFlect \citep{DBLP:journals/corr/abs-2411-14721} & \uline{0.886} 
 & 0.430  & 13.99  & \uline{0.916}  & 0.828  & 0.775  & \uline{0.981} \\
\rowcolor{lightblue}
MolBridge-Gen-small & 0.827 & 0.266 & 16.88 & 0.898 & 0.820 & 0.751 & 0.947 \\ 
\rowcolor{lightblue}
MolBridge-Gen-base & 0.842 & \uline{0.358} & 15.66 & \textbf{0.918} & \textbf{0.854} & \textbf{0.798} & 0.956  \\
\bottomrule
\end{tabular}%
}
\caption{Results of molecule generation task on CheBI-20 test set. \textbf{Bold} and \uline{underlined} indicate the best and second-best results, respectively. Full comparison is in Table \ref{tab:molgen_appendix} in the Appendix.}
\label{tab:molgen}
\end{table*}

\subsection{Molecular Property Prediction}
\label{property}
\paragraph{Settings.}
Following \citet{zhang2025atomas}, we evaluate MolBridge on eight classification datasets from MoleculeNet. We use the scaffold split provided by DeepChem \citep{Ramsundar-et-al-2019}, and we report the ROC-AUC scores. We jointly train the MolBridge molecule encoder and text encoder to see whether our proposed framework empowers the fine-grained understanding of molecules. We compare our model with multimodal methods.

\paragraph{Results.} 
Table~\ref{tab:property} shows the results of eight property prediction tasks.
Although MolBridge is much smaller than the previous fine-grained alignment method, it achieves a 2.2\%p improvement in performance.
This demonstrates that the substructural relation–based alignment approach enhances MolBridge’s ability to capture fine-grained structural information.
Because property prediction often requires distinguishing subtle differences between similar molecules \citep{DBLP:conf/emnlp/ParkPKL024}, the performance gain suggests that our method enables the model to learn precise substructural representations, thereby capturing nuanced relationships and effectively transferring molecular knowledge to a wide range of prediction tasks.

\begin{figure*}[ht!]
    \centering
    \includegraphics[width=\linewidth]{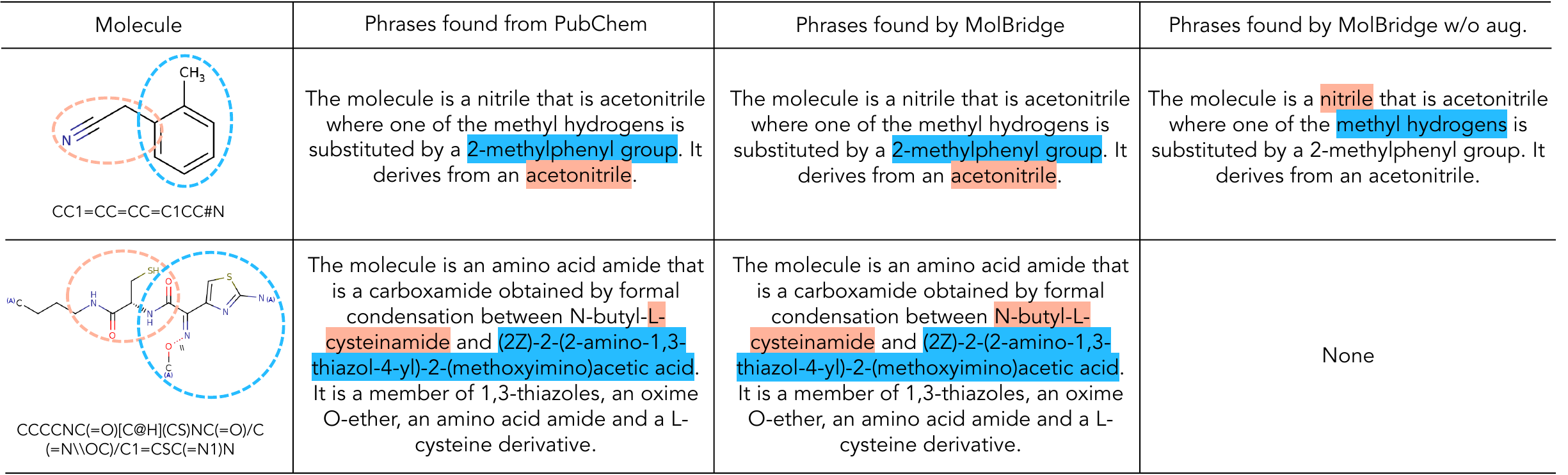}
    \caption{Examples of chemical phrase retrieval. Identified substructures (dashed circles) and retrieved phrases from PubChem, MolBridge, and MolBridge w/o augmentation.}
    \label{fig: keyword_retrieved}
    \vspace{-0.5cm}
\end{figure*}

\begin{figure}[t!]
    \centering
    \begin{subfigure}[t]{0.5\columnwidth}
        \centering
        \includegraphics[width=\columnwidth]{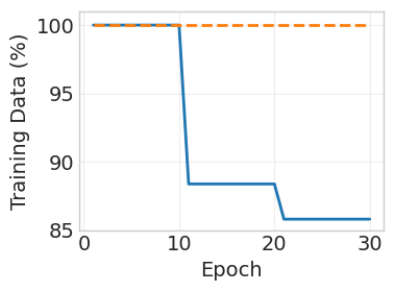}
        \caption{Percentage of train data.}
    \end{subfigure}%
    \begin{subfigure}[t]{0.5\columnwidth}
        \centering
        \includegraphics[width=\columnwidth]{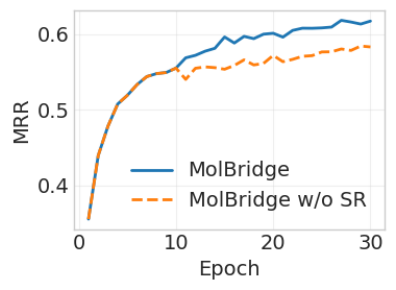}
        \caption{Performance (MRR; M2T).}
    \end{subfigure}
    \caption{Analysis of the impact of Self-Refinement (SR). We evaluate both MolBridge and MolBridge trained without SR on PCDes scaffold test set.}
    \label{fig: noisy}
\end{figure}

\subsection{Molecule Captioning \& Generation}
\label{captioning}
\paragraph{Settings.}
We evaluate MolBridge-Gen on molecule captioning and generation tasks using the CheBI-20 dataset. To assess the quality of generated captions, we use evaluation metrics that include BLEU \citep{bleu}, ROUGE \citep{lin-2004-rouge}, and METEOR \citep{meteor} scores.
For the de novo molecule generation task, we employ BLEU to measure the percentage of predictions that exactly match the true labels (Exact Match; EM), Levenshtein distance \citep{miller2009levenshtein} for string similarity, Validity for grammatical correctness of the generated molecules, and molecular fingerprint-based similarity measures such as MACCS FTS \citep{DBLP:journals/jcisd/DurantLHN02}, RDK FTS \citep{DBLP:journals/jcisd/SchneiderSL15}, and Morgan FTS \citep{DBLP:journals/jcisd/RogersH10} to compare similarity with the original molecules.\footnote{Additional experimental results on PubChem324k are in Appendix \ref{app: 324k}.}

\paragraph{Results.}
Tables \ref{tab:molcap} and \ref{tab:molgen} present the results for molecule captioning and generation. Despite being based on MolT5-small/base, both MolBridge-Gen-small/base show improvements over MolT5-large and other baselines.
In the captioning task, MolBridge-Gen-base achieves the highest scores in ROUGE scores and METEOR, outperforming all baseline models, including the much larger 7B MolReFlect. This suggests that learning substructure–phrase relationships enables more fine-grained understanding of molecular content, even with smaller models. Moreover, these improvements demonstrate that the substructure–phrase pairs identified by MolBridge for training MolBridge-Gen are indeed effective in learning fine-grained alignments between molecules and textual descriptions, thereby enhancing the model’s generative capabilities.\footnote{More analysis of molecule captioning is in Appendix~\ref{app: molcap}.}

In the molecule generation task, MolBridge-Gen-base achieves the best fingerprint-based similarity scores across MACCS, RDK , and Morgan metrics, and outperforms Atomas-base, which conducts an implicit fine-grained alignment, while reaching comparable performance to Atomas-large. This indicates that explicitly modeling local structure–text relationships enables the model to generate molecules that are semantically aligned with input descriptions.
These suggest that our explicit fine-grained alignment strategy enhances the model’s capacity to encode and decode chemically meaningful information, leading to improved performance in both captioning and generation tasks, even with smaller sizes.\footnote{Generated examples are shown in Figure \ref{fig:cap_ex} and \ref{fig:mol_ex}.}

\section{Analysis}
In this section, we analyze the fine-grained alignment capability of MolBridge. Further analyses on model choice, ablation study, and error analysis on the molecular scale and complexity are provided in Appendix~\ref{choice}, \ref{ablation}, and \ref{app:error}.

\paragraph{Case study.}
To investigate whether the local relations discovered by MolBridge accurately capture meaningful substructure–phrase correspondences, we conduct a qualitative evaluation. Figure~\ref{fig: keyword_retrieved} illustrates representative examples comparing retrieved chemical phrases from MolBridge, a baseline model trained without augmentation, and ground-truth phrases curated from the PubChem database. Each molecule is annotated with dashed circles indicating identified substructures, and the corresponding phrases are highlighted with matching colors.
In both cases, MolBridge retrieves phrases that closely align with the ground truth, demonstrating its ability to identify valid local relationships. In contrast, the MolBridge trained without augmentation retrieves either irrelevant or no phrases. These results suggest that substructure-aware alignments enables the model to learn more precise and semantically meaningful mappings between molecular and textual fragments.

\paragraph{Impact of self-refinement.}
\label{analysis: refinement}
To assess the effectiveness of the self-refinement process in MolBridge, we compare retrieval performance between models trained with and without refinement, as shown in Figure~\ref{fig: noisy}. We observe that the refined model consistently outperforms the baseline over training epochs, despite filtering out approximately 15\% of the training data in two stages.
The majority of the removed pairs are substructure–caption relations, particularly cases where the entire molecule is incorrectly treated as a substructure during decomposition. As illustrated in Figure~\ref{fig:filtered} in the Appendix, such cases include molecules misidentified as their own substructures, leading to invalid substructural relations, as well as overly generic captions that fail to capture substructure-level semantics.
These results indicate that removing noisy supervision signals during training helps construct a cleaner and more informative dataset, ultimately improving model robustness and alignment quality.

\begin{table}[t]
\footnotesize
\centering
\begin{tabular}{@{}c|cc|cc@{}}
\toprule
\multirow{2}{*}{Phrases Extractor} & \multicolumn{2}{c|}{T2M} & \multicolumn{2}{c}{M2T} \\
 & R@1 & MRR & R@1 & MRR \\ \midrule
ChemDataExtractor & \textbf{34.38} & \textbf{45.73} & \textbf{36.45} & \textbf{47.82} \\
GPT-4 + MolT5 & 23.89 & 34.99 & 27.33 & 38.88 \\ \bottomrule
\end{tabular}%
\caption{Evaluation results on PCDes scaffold test set with different phrase extractors for MolBridge.}
\label{tab:keyword}
\end{table}

\begin{table}[t]
\footnotesize
\centering
\begin{tabular}{@{}c|cc|cc@{}}
\toprule
\multirow{2}{*}{Decompose Method} & \multicolumn{2}{c|}{T2M} & \multicolumn{2}{c}{M2T} \\
 & R@1 & MRR & R@1 & MRR \\ \midrule
BRICS & \textbf{34.38} & \textbf{45.73} & \textbf{36.45} & \textbf{47.82} \\
RECAP & 19.35 & 30.27 & 21.05 & 32.53 \\ \bottomrule
\end{tabular}%
\caption{Evaluation results on PCDes scaffold test set with different substructure extractors for MolBridge.}
\label{tab:rdkit}
\end{table}



\paragraph{Analysis on Fragment Extractor}\label{extractor}
In our search for optimal tools to extract substructures and phrases, we employ various techniques and analyze their differences by training MolBridge for three epochs. The retrieval performances are reported in Table \ref{tab:keyword} and \ref{tab:rdkit}. 

For molecular decomposition, we use RECAP \citep{degen2008art} and BRICS \citep{lewell1998recap}. The key difference between them lies in the number of bond types considered, with BRICS enabling MolBridge to explore a broader range of substructures and yield more positive results.  
For phrase extraction, ChemDataExtractor (CDE) often produces incomplete outputs, so we design an LLM-based extractor \citep{molregpt}. Using GPT-4 \citep{DBLP:journals/corr/abs-2303-08774} on 10K sampled captions and fine-tuning MolT5-large, we expand phrase coverage. However, this reduces diversity, making CDE comparatively more effective. These findings underscore the importance of collecting diverse substructures and phrases for fine-grained alignment.

\begin{table}[t]
\centering
\footnotesize
\begin{tabular}{@{}l|c@{}}
\toprule
\multicolumn{1}{c|}{Method} & Ranking$\downarrow$ \\ \midrule
MolT5-large \citep{molt5}                & 2.4 (0/3/2)   \\
Atomas-base \citep{zhang2025atomas}      & 2.2 (1/2/2)    \\
MolBridge-Gen-base         & \textbf{1.4} (4/0/1)   \\ \bottomrule
\end{tabular}
\caption{Human evaluation results. Ranking refers to the average ranking of human evaluation. The numbers in brackets indicate the counts of ranks 1, 2, and 3.}
\label{tab:human_eval}
\end{table}

\paragraph{Human evaluation}

Following the human evaluation settings of \citet{zhang2025atomas}, we conduct experiments in which annotators are asked to rank five randomly selected captions for each method according to their closeness to the ground truth. Specifically, we compare MolBridge-Gen-base against MolT5-large and Atomas-base with three human annotators. Table~\ref{tab:human_eval} reports the average rankings. MolBridge-Gen-base achieves the highest average ranking, placing first in 4 out of 5 generated captions in our human evaluation. More details, including evaluation instructions and generated examples, are provided in Appendix~\ref{app: human_eval}.

\section{Conclusion}
We have presented MolBridge, a substructure-aware framework for fine-grained molecule–text alignment. By explicitly aligning molecular substructures with corresponding chemical phrases, MolBridge enables fragment-level representation learning that better captures subtle differences between molecules and their descriptions.
Our experiments show that MolBridge consistently outperforms existing models across retrieval, property prediction, and generation tasks, demonstrating the effectiveness of leveraging localized alignment signals in multimodal molecular learning.

\section*{Limitations}
While MolBridge demonstrates strong performance across multiple molecule–text tasks, several limitations remain.

\paragraph{Reliance on fragment extractors.}
MolBridge relies on external extractors to identify molecular substructures and chemical phrases for alignment. To address potential noise and inaccuracies introduced during this process, we incorporate a self-refinement mechanism that filters unreliable alignment signals and contributes meaningfully to the robustness of MolBridge. Even with this mechanism, the alignment quality is still influenced by the choice of extractor, as can be seen in Appendix \ref{extractor}. Developing more customized or domain-specific extractors could further improve the precision of fragment-level alignment in future work.

\paragraph{Limited structural exploration.}
MolBridge operates solely on 1D SMILES representations yet consistently outperforms models that directly utilize 2D molecular graphs. This result highlights the effectiveness of our fragment-level alignment approach. Nevertheless, incorporating additional structural information such as 2D topology or 3D conformations could provide complementary benefits \citep{molca, DBLP:conf/bibm/XiaoCZH24}. Extending MolBridge in this direction may further enhance its ability to model complex spatial relationships in molecular data.

\section*{Acknowledgments}
This work was supported by the National Research Foundation of Korea (NRF) grant funded by the Korea government (MSIT) (No.RS-2025-00517221 and No.RS-2024-00415812) and Institute of Information \& communications Technology Planning \& Evaluation (IITP) grant funded by the Korea government (MSIT) (No.RS-2024-00439328, Karma: Towards Knowledge Augmentation for Complex Reasoning (SW Starlab), No.RS-2024-00457882, AI Research Hub Project, and No.RS-2019-II190079, Artificial Intelligence Graduate School Program (Korea University)).

\bibliography{custom}

\clearpage
\appendix

\begin{center}
\large
\textbf{Appendix}    
\end{center}

\section{Implementation Details} 
\label{implementation}
We initialize MolBridge with two specialized encoders: MoLFormer-XL \citep{molformer}\footnote{\href{https://huggingface.co/ibm-research/MoLFormer-XL-both-10pct}{ibm-research/MoLFormer-XL-both-10pct}} for SMILES input and SciBERT \citep{scibert}\footnote{\href{https://huggingface.co/allenai/scibert\_scivocab\_uncased}{allenai/scibert\_scivocab\_uncased}} for textual input. Noisy supervision signals are filtered out every 10 epochs.
We explored learning rates from the set \{1e-5, 3e-5, 1e-4, 5e-4\} and batch sizes from \{32, 64, 128, 256, 512\}, and report results from the best-performing configurations. The sequence length for both SMILES and text inputs is fixed to 256 tokens. For MolBridge, we use a learning rate of 2e-4 and a batch size of 256 during pretraining. For MolBridge-Gen, which is built upon MolT5 \citep{molt5}, we pre-train the model with a learning rate of 5e-4 and fine-tune it with 1e-4. A batch size of 128 is used for both pretraining and fine-tuning of MolBridge-Gen.

For molecular property prediction using MolBridge trained over 10 epochs, we conducted a full grid search over the learning rates and batch sizes mentioned above to select the best-performing configuration for each task.
All models are trained for 50 epochs using canonicalized SMILES and the AdamW optimizer. We apply gradient accumulation to handle large batch sizes across four NVIDIA A5000 GPUs.
To extract structure–phrase pairs for generative training, we empirically set the MolBridge score threshold $\tau$ to 0.3. For retrieval tasks, we adopt a zero-shot setting to ensure a fair comparison with previous works \citep{molca, zhang2025atomas}.

\section{Analysis on Model Choice}
\label{choice}
We aim to assess the impact of molecular and scientific literature understanding on the encoders used as the backbone of MolBridge. To this end, we initialize the model with ChemBERTa \citep{chemberta}, which has been reported to show lower molecular property prediction performance compared to MoLFormer-XL, and with BERT \citep{bert}, trained on general domain texts. We train the models for three epochs, as these results show a similar tendency to final results under the same training objectives. 
The retrieval results on the PCDes scaffold test set in Table~\ref{tab:model_choice} show decreases of 20\%p and 12.8\%p in MRR for text-to-molecule and molecule-to-text retrieval, respectively.
This underscores the critical role of molecular understanding and scientific literature comprehension in molecule and text alignment.

\begin{table}[h]
\centering
\resizebox{\columnwidth}{!}{%
\begin{tabular}{@{}cc|cc|cc@{}}
\toprule
SMILES & Text & \multicolumn{2}{c|}{T2M} & \multicolumn{2}{c}{M2T} \\
Encoder & Encoder & R@1 & MRR & R@1 & MRR \\ \midrule
MoLFormer-XL & SciBERT & \textbf{34.38} & \textbf{45.73} & \textbf{36.45} & \textbf{47.82} \\
ChemBERTa & BERT & 15.57 & 25.73 & 23.42 & 34.98 \\ \bottomrule
\end{tabular}
}
\caption{Evaluation results on PCDes scaffold test set with different model choices for MolBridge.}
\label{tab:model_choice}
\end{table}

\begin{table*}[ht]
\small
\centering
\begin{tabular}{@{}l|cccc|cccc@{}}
\toprule
\multirow{2}{*}{Methods} & \multicolumn{4}{c|}{Text to Molecule} & \multicolumn{4}{c}{Molecule to Text} \\
 & R@1 & R@5 & R@10 & MRR & R@1 & R@5 & R@10 & MRR \\ \midrule
MolBridge & \textbf{34.38} & 58.84 & 67.86 & 45.73 & \textbf{36.45} & \textbf{61.38} & \textbf{70.33} & \textbf{47.82} \\ 
MolBridge w/o type classification & 34.31 & \textbf{59.87} & \textbf{68.23} & \textbf{46.01} & 35.12 & 60.57 & 69.23 & 46.93 \\ 
MolBridge w/o multi-positive contrastive learning & 29.27 & 54.73 & 63.85 & 41.03 & 32.14 & 58.30 & 67.56 & 43.96 \\
MolBridge w/o substructure-caption pairs & 23.92 & 49.42 & 60.11 & 35.83 & 26.23 & 51.19 & 61.68 & 38.02 \\ 
MolBridge w/o molecule-phrase pairs & 19.48 & 42.43 & 52.62 & 30.55 & 22.22 & 48.51 & 59.51 & 34.51 \\
MolBridge w/o augmentation & 13.97 & 34.65 & 45.24 & 23.96 & 14.73 & 37.82 & 49.48 & 25.96 \\ \bottomrule
\end{tabular}
\caption{Ablation study of MolBridge on PCDes scaffold test set. Each model is trained over 3 epochs, as we observed that these results show a similar tendency to the final results.}
\label{tab:molbridge_ablation}
\end{table*}

\begin{table*}[ht!]
\small
\centering
\resizebox{\textwidth}{!}{%
\begin{tabular}{@{}l|c|c|cccccc@{}}
\toprule
Method & $\tau$ & \# Pairs & BLEU-2$\uparrow$ & BLUE-4$\uparrow$ & ROUGE-1$\uparrow$ & ROUGE-2$\uparrow$ & ROUGE-L$\uparrow$ & METEOR$\uparrow$ \\ \midrule
MolT5-base & - & - & 0.549 & 0.457 & 0.635 & 0.481 & 0.576 & 0.580 \\
Atomas-base & - & - & 0.632 & 0.545 & 0.686 & 0.545 & 0.626 & - \\
MolBridge-Gen-base & 0.2 & 163k & 0.629 & 0.547 & 0.687 & 0.551 & 0.631 & 0.651 \\
MolBridge-Gen-base & 0.3 & 32k & \textbf{0.674} & \textbf{0.605} & \textbf{0.724} & \textbf{0.609} & \textbf{0.676} & \textbf{0.693} \\
MolBridge-Gen-base & 0.4 & 5k & 0.543 & 0.447 & 0.619 & 0.463 & 0.560 & 0.567 \\ \bottomrule
\end{tabular}
}
\caption{Molecule captioning performance of MolBridge-Gen with different cosine similarity thresholds on ChEBI-20.}
\label{tab:threhold_cap}
\end{table*}

\begin{table*}[ht!]
\small
\centering
\resizebox{\textwidth}{!}{%
\begin{tabular}{@{}l|c|c|ccccccc@{}}
\toprule
Method & $\tau$ & \# Pairs & BLEU$\uparrow$ & EM$\uparrow$ & Levenshtein$\downarrow$ & MACCS FTS$\uparrow$ & RDK FTS$\uparrow$ & Morgan FTS$\uparrow$ & Validity$\uparrow$ \\ \midrule
MolT5-base & - & - & 0.854 & 0.318 & 16.32 & 0.889 & 0.813 & 0.750 & 0.958 \\
Atomas-base & - & - & \textbf{0.868} & 0.343 & \textbf{13.76} & 0.908 & 0.827 & 0.773 & 0.971 \\ 
MolBridge-Gen-base & 0.2 & 163k & 0.834 & 0.278 & 16.11 & 0.901 & 0.822 & 0.758 & 0.956 \\ 
MolBridge-Gen-base & 0.3 & 32k & 0.842 & \textbf{0.358} & 15.66 & \textbf{0.918} & \textbf{0.854} & \textbf{0.798} & 0.956 \\ 
MolBridge-Gen-base & 0.4 & 5k & 0.783 & 0.173 & 21.82 & 0.853 & 0.750 & 0.670 & 0.943 \\ \bottomrule
\end{tabular}
}
\caption{Molecule generation performance of MolBridge-Gen with different cosine similarity thresholds on ChEBI-20.}
\label{tab:threhold_mol}
\end{table*}

\section{Ablation study}
\label{ablation}
To validate the effectiveness of each component in the MolBridge framework, we conduct an ablation study by training the same model for three epochs under different objective configurations and augmentation settings. The results, summarized in Table~\ref{tab:molbridge_ablation}, show that each proposed component contributes meaningfully to overall performance.

First, we observe that removing our augmentation strategy causes a dramatic drop in retrieval performance, with an average decrease of 21.8\%p in MRR compared to the full model. This highlights the importance of explicitly modeling substructural relationships for fine-grained alignment.

When we remove either the substructure-caption or the molecule-phrase pairs, the performance still improves relative to the MolBridge w/o augmentation. This indicates that even partial fragment-level supervision is beneficial. Among the two, the absence of molecule-phrase alignment leads to a larger drop, which suggests that identifying diverse and accurate phrases plays an important role in learning meaningful semantic correspondences.

We also find that removing multi-positive contrastive learning leads to a 4.28\%p decrease in average MRR. This result supports the assumption that a single molecule or caption can correspond to multiple relevant fragments, and confirms that the proposed objective effectively captures such compositional relationships.

Lastly, we examine the effect of removing the type classification loss used during the self-refinement process. The results show a slight drop in molecule-to-text retrieval performance, although a marginal improvement is observed in the reverse direction. Despite this, the overall benefit of the self-refinement mechanism remains clear, as it enables the model to filter noisy alignment signals during training and contributes to the stability and robustness of representations, as discussed in Section \ref{analysis: refinement}.

\section{Analysis on Cosine Similarity Threshold}
\label{app: threshold}
We evaluate the effect of different threshold values ($\tau$) on the generative performance of our model. Table \ref{tab:threhold_cap} and \ref{tab:threhold_mol} show the results of pre-training and fine-tuning MolBridge-Gen on the ChEBI-20 dataset using three different thresholds. All models were trained for the same number of steps to ensure a fair comparison.

Our experiments indicate that a threshold of 0.3 yields the best overall performance. When the threshold is set to 0.2, MolBridge-Gen achieves results comparable to the Atomas, suggesting our approach remains robust. However, at a threshold of 0.4, performance drops to the level of, or slightly below, the backbone of our model (MolT5). We attribute this to the substantial reduction in the number of original pairs containing fragment matches, leading to overfitting due to limited training data.

\begin{table*}[ht]
\small
\centering
\resizebox{\textwidth}{!}{%
\begin{tabular}{@{}lccccccc@{}}
\toprule
\multicolumn{1}{l|}{Method} & \multicolumn{1}{c|}{\# Pairs} & BLEU-2$\uparrow$ & BLUE-4$\uparrow$ & ROUGE-1$\uparrow$ & ROUGE-2$\uparrow$ & ROUGE-L$\uparrow$ & METEOR$\uparrow$ \\ \midrule
\multicolumn{1}{l|}{MolT5-base} & \multicolumn{1}{c|}{-} & 0.549 & 0.457 & 0.635 & 0.481 & 0.576 & 0.580 \\ \midrule
\multicolumn{1}{l|}{MolT5-base + inital training} & \multicolumn{1}{c|}{432k} & 0.580 & 0.495 & 0.657 & 0.516 & 0.600 & 0.611 \\  
\multicolumn{1}{l|}{MolBridge-Gen-base} & \multicolumn{1}{c|}{32k} & \textbf{0.674} & \textbf{0.605} & \textbf{0.724} & \textbf{0.609} & \textbf{0.676} & \textbf{0.693} \\  
\bottomrule
\end{tabular}%
}
\caption{Analysis on pre-training approach of MolBridge-Gen on CheBI-20 test set. Initial training refers to the training of the model before fine-tuning with our curated dataset without augmentation described in Section \ref{method:expansion}.}
\label{tab:gen-pretrain}
\end{table*}
\section{Analysis on Pretraining Strategy for Generation}
\label{app: molcap}
We investigate the effect of our pretraining strategy on molecule captioning by comparing MolBridge-Gen with its baseline, as shown in Table \ref{tab:gen-pretrain}. To this end, we fine-tune both the original MolT5-base and the MolT5-base pretrained on our curated dataset without augmentation.

\begin{figure*}[ht]
\centering
  \includegraphics[width=0.7\linewidth]{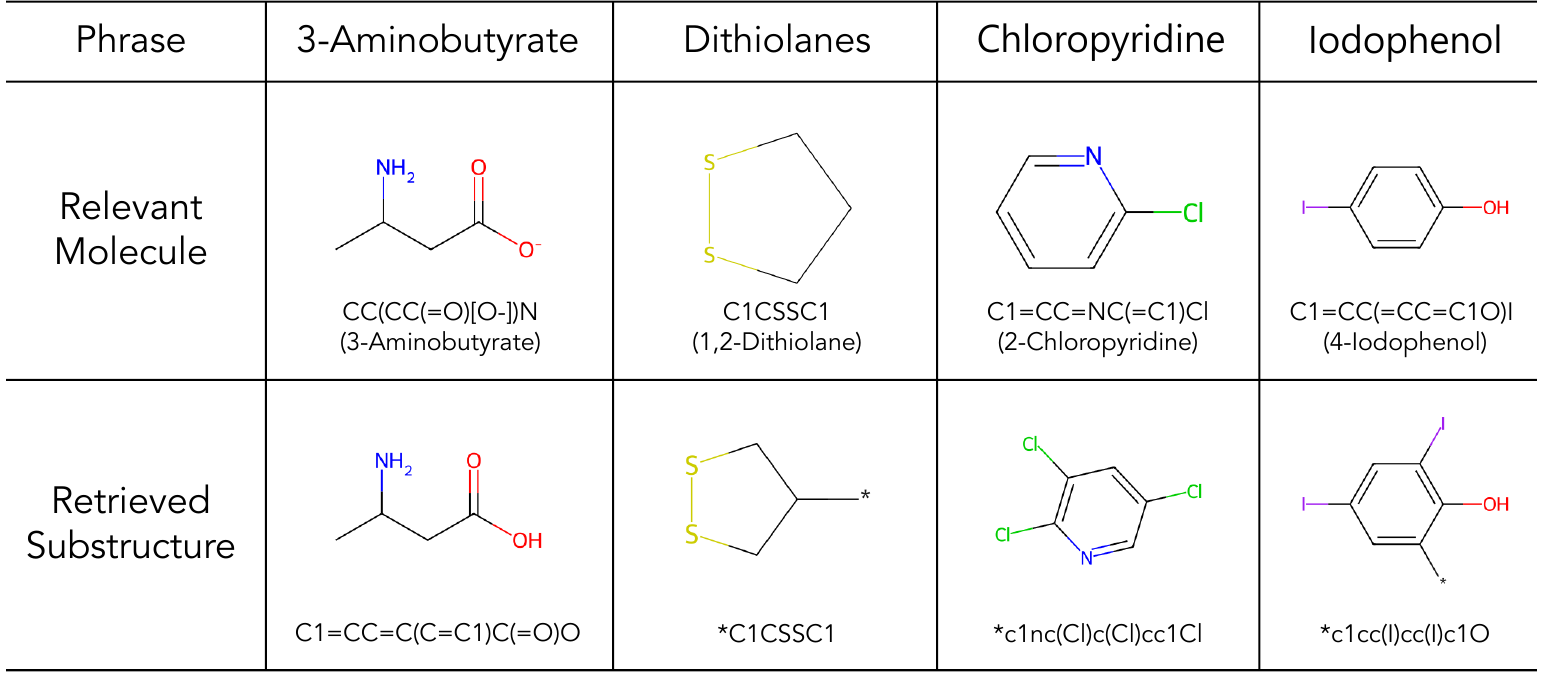}
    \caption{Retrieved substructures using chemical phrase by MolBridge along with the phrase-relevant molecules found in PubChem.}
    \label{fig:sub}
\end{figure*}

The results in Table \ref{tab:gen-pretrain} indicate that initial pretraining enhances generation quality, yielding consistent improvements across all evaluation metrics. More importantly, MolBridge-Gen, which is trained on only 32k molecule–caption pairs augmented with local alignment signals discovered by MolBridge (Figure \ref{fig:sub}), surpasses all other settings by a substantial margin. In particular, it achieves gains of 9.4 to 11.0 points in BLEU-2, BLEU-4, and METEOR scores over the original MolT5 model, despite using far fewer training examples. These findings demonstrate that the local structure–language correspondences captured by MolBridge provide more informative supervision than large-scale pretraining without alignment, underscoring the advantage of explicitly modeling fine-grained relationships for molecule captioning.

\begin{table}[t!]
\centering
\resizebox{\columnwidth}{!}{%
\begin{tabular}{@{}l|cc|cc@{}}
\toprule
\multirow{2}{*}{Methods} & \multicolumn{2}{c|}{T2M} & \multicolumn{2}{c}{M2T} \\
 & R@1 & R@20 & R@1 & R@20 \\ \midrule
\rowcolor{gray}
\multicolumn{5}{l}{1D SMILES + 2D Graph} \\
MoMu-S \citep{momu} & - & 75.5 & - & 79.1 \\
MoMu-K \citep{momu} & - & 79.0 & - & 80.2 \\
MoleculeSTM \citep{moleculeSTM} & 35.8 & 77.0 & 39.5 & 80.4 \\
MolCA \citep{molca} & 46.0 & 82.3 & 48.1 & 85.6 \\ \midrule
\rowcolor{gray}
\multicolumn{5}{l}{1D SMILES} \\
SciBERT \citep{scibert} & - & 60.8 & - & 60.7 \\
KV-PLM \citep{kvplm}& - & 64.3 & - & 75.9 \\
\rowcolor{lightblue}
MolBridge & \textbf{54.2} & \textbf{86.7} & \textbf{57.0} & \textbf{88.6} \\ \bottomrule
\end{tabular}
}
\caption{Zero-shot molecule-text retrieval performance on PCDes test set. The results of baselines are borrowed from \citep{molca}.}
\label{tab:ret3}
\end{table}

\section{Error Analysis on Molecular Scale and Complexity}
\label{app:error}

To better understand the behavior of MolBridge-Gen, we conduct an error analysis on the ChEBI-20 dataset by dividing the test set according to the median values of molecular scale (atom count) and molecular complexity (BertzCT \cite{bertz1981first}). 
We evaluate our model on low/high scale and low/high complexity subsets for both molecule captioning and molecule generation tasks. 
Results are shown in Table \ref{tab:chebi_scale_caption}, \ref{tab:chebi_complexity_caption}, \ref{tab:chebi_scale_generation}, and \ref{tab:chebi_complexity_generation}.
We find that MolBridge-Gen makes more errors on molecules with lower scale and lower complexity, while performance is higher for larger and more complex molecules. 
This suggests that complex and large molecules contain richer compositional relationships between substructures and language, which our model is designed to capture.

\begin{table*}[th]
\centering
\footnotesize
\begin{tabular}{l|cccccc}
\toprule
MolBridge-Gen-base & BLEU-2$\uparrow$ & BLEU-4$\uparrow$ & ROUGE-1$\uparrow$ & ROUGE-2$\uparrow$ & ROUGE-L$\uparrow$ & METEOR$\uparrow$ \\
\midrule
Low scale set   & 0.660 & 0.589 & 0.715 & 0.598 & 0.667 & 0.683 \\
High scale set  & \textbf{0.687} & \textbf{0.620} & \textbf{0.735} & \textbf{0.622} & \textbf{0.687} & \textbf{0.704} \\
Original set    & 0.674 & 0.605 & 0.724 & 0.609 & 0.676 & 0.693 \\
\bottomrule
\end{tabular}
\caption{Evaluation results on the ChEBI-20 test set with respect to molecular scale (Molecule Captioning).}
\label{tab:chebi_scale_caption}
\end{table*}

\begin{table*}[th]
\centering
\footnotesize
\begin{tabular}{l|cccccc}
\toprule
MolBridge-Gen-base & BLEU-2$\uparrow$ & BLEU-4$\uparrow$ & ROUGE-1$\uparrow$ & ROUGE-2$\uparrow$ & ROUGE-L$\uparrow$ & METEOR$\uparrow$ \\
\midrule
Low complexity set & 0.667 & 0.595 & 0.717 & 0.599 & 0.670 & 0.686 \\
High complexity set& \textbf{0.680} & \textbf{0.614} & \textbf{0.732} & \textbf{0.620} & \textbf{0.683} & \textbf{0.701} \\
Original set       & 0.674 & 0.605 & 0.724 & 0.609 & 0.676 & 0.693 \\
\bottomrule
\end{tabular}
\caption{Evaluation results on the ChEBI-20 test set with respect to molecular complexity (Molecule Captioning).}
\label{tab:chebi_complexity_caption}
\end{table*}

\begin{table*}[th]
\centering
\footnotesize
\begin{tabular}{l|cccccccc}
\toprule
MolBridge-Gen-base & BLEU$\uparrow$ & EM$\uparrow$ & Levenshtein$\downarrow$ & MACCS FTS$\uparrow$ & RDK FTS$\uparrow$ & Morgan FTS$\uparrow$ & Validity$\uparrow$ \\
\midrule
Low scale set   & \textbf{0.844} & \textbf{0.438} & \textbf{6.538}  & 0.900 & 0.822 & 0.777 & \textbf{0.988} \\
High scale set  & 0.831 & 0.271 & 25.506 & \textbf{0.938} & \textbf{0.892} & \textbf{0.821} & 0.921 \\
Original set    & 0.842 & 0.358 & 15.660 & 0.918 & 0.854 & 0.798 & 0.956 \\
\bottomrule
\end{tabular}
\caption{Evaluation results on the ChEBI-20 test set with respect to molecular scale (Molecule Generation).}
\label{tab:chebi_scale_generation}
\end{table*}

\begin{table*}[th]
\centering
\footnotesize
\begin{tabular}{l|cccccccc}
\toprule
MolBridge-Gen-base & BLEU$\uparrow$ & EM$\uparrow$ & Levenshtein$\downarrow$ & MACCS FTS$\uparrow$ & RDK FTS$\uparrow$ & Morgan FTS$\uparrow$ & Validity$\uparrow$ \\
\midrule
Low complexity set & \textbf{0.860} & \textbf{0.433} & \textbf{6.382}  & 0.909 & 0.837 & 0.795 & \textbf{0.992} \\
High complexity set& 0.827 & 0.282 & 24.948 & \textbf{0.927 }& \textbf{0.873} & \textbf{0.801} & 0.919 \\
Original set       & 0.842 & 0.358 & 15.660 & 0.918 & 0.854 & 0.798 & 0.956 \\
\bottomrule
\end{tabular}
\caption{Evaluation results on the ChEBI-20 test set with respect to molecular complexity (Molecule Generation).}
\label{tab:chebi_complexity_generation}
\end{table*}

\section{Evaluation on PubChem324k}
\label{app: 324k}

To further assess the generalizability of MolBridge-Gen beyond ChEBI-20, we evaluate the model on the PubChem324k dataset \citep{molca} for both molecule captioning and molecule generation tasks. 
As shown in Tables~\ref{tab:pubchem_captioning} and \ref{tab:pubchem_generation}, 
MolBridge-Gen achieves strong and consistent performance across different models and evaluation metrics.

\section{Details of Human Evaluation}
\label{app: human_eval}
\begin{figure*}[ht]
    \centering
    \includegraphics[width=\linewidth]{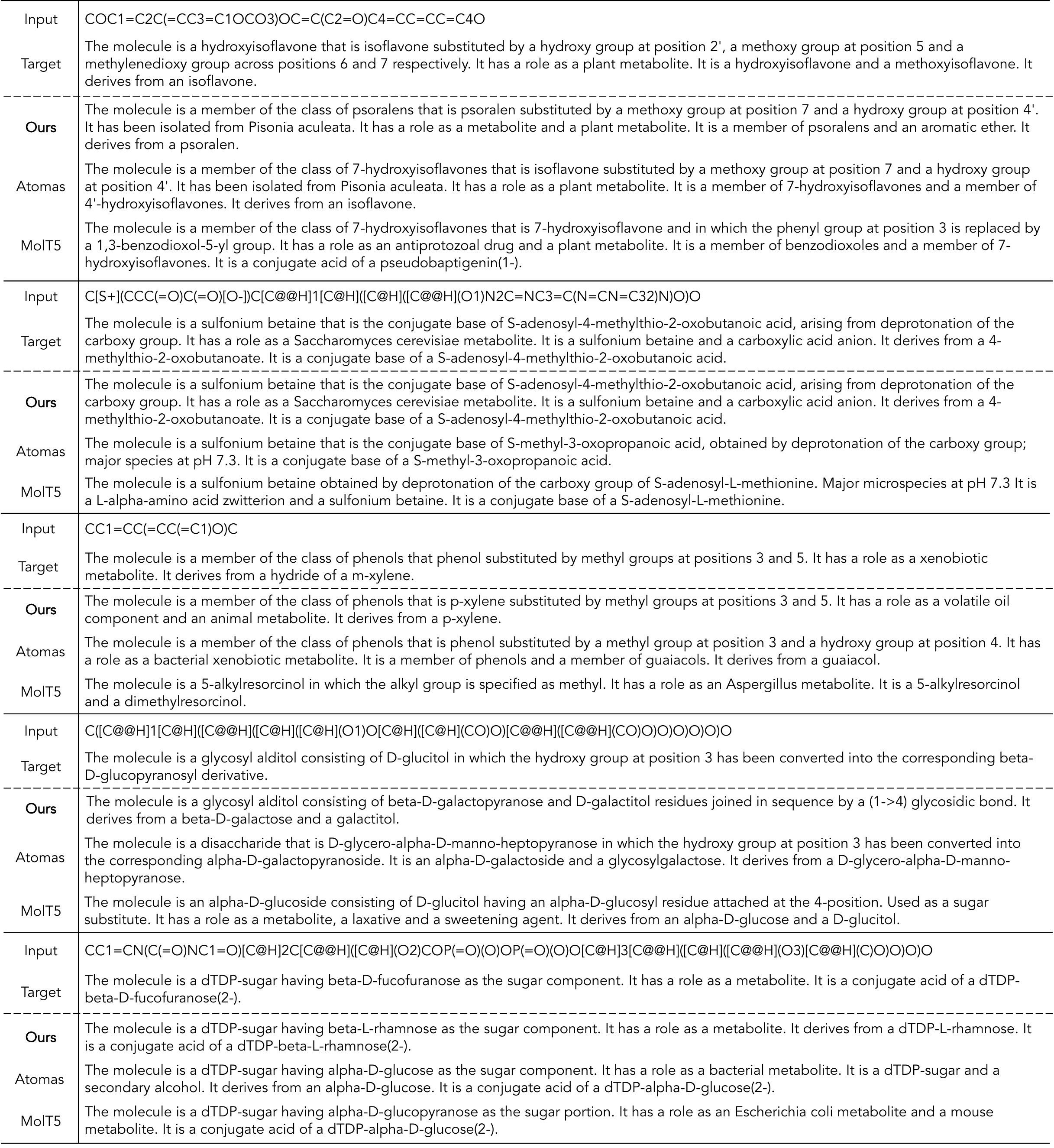}
    \caption{Examples used for human evaluation.}
    \label{fig:human_eval}
\end{figure*}

We invited three NLP experts as annotators, specifically those with prior publications or project experience in related domains. The annotators participated on a voluntary basis, and no payment was provided. For the evaluation, we randomly sampled five molecules. For each molecule, annotators were given the molecule structure, its ground-truth caption, and three generated captions produced by MolT5-large, Atomas-base, and MolBridge-Gen-base. The generated captions were presented in randomized order, and annotators were instructed to “rank the three models (Model~1, Model~2, Model~3) according to their relevance to the ground-truth caption.” The five evaluation examples used in this study are shown in Figure~\ref{fig:human_eval}.

\begin{table*}[t]
\centering
\small
\resizebox{\textwidth}{!}{%
\begin{tabular}{l|c|cccccc}
\toprule
Method & \#Params & BLEU-2$\uparrow$ & BLEU-4$\uparrow$ & ROUGE-1$\uparrow$ & ROUGE-2$\uparrow$ & ROUGE-L$\uparrow$ & METEOR$\uparrow$ \\
\midrule
MolT5-base \citep{molt5} & 248M & 0.301 & 0.209 & 0.403 & 0.251 & 0.338 & 0.356 \\
MolCA-Galactica-1.3B \cite{molca} & 1.3B & 0.387 & 0.303 & 0.502 & 0.359 & 0.445 & 0.456 \\
ICMA-Mistral-7B \cite{DBLP:journals/corr/abs-2403-04197_icma} & 7B & 0.416 & 0.345 & 0.505 & 0.367 & 0.453 & 0.464 \\
MolBridge-Gen-base & 248M & \textbf{0.479} & \textbf{0.421} & \textbf{0.596} & \textbf{0.486} & \textbf{0.554} & \textbf{0.549} \\
\bottomrule
\end{tabular}
}
\caption{Molecule captioning results on the PubChem324k dataset.}
\label{tab:pubchem_captioning}
\end{table*}

\begin{table*}[t]
\centering
\small
\resizebox{\textwidth}{!}{%
\begin{tabular}{l|c|ccccccc}
\toprule
Method & \#Params & BLEU$\uparrow$ & EM$\uparrow$ & Levenshtein$\downarrow$ & MACCS FTS$\uparrow$ & RDK FTS$\uparrow$ & Morgan FTS$\uparrow$ & Validity$\uparrow$ \\
\midrule
ICMA-Mistral-7B \cite{DBLP:journals/corr/abs-2403-04197_icma} & 7B   & 0.526 & 0.163 & 62.25  & 0.799 & 0.678 & 0.573 & 0.935 \\
Atomas-large \cite{zhang2025atomas}    & 825M & 0.734 & --    & 28.186 & 0.773 & 0.637 & 0.535 & 0.945 \\
MolBridge-Gen-base & 248M & \textbf{0.742} & \textbf{0.200} & \textbf{27.294} & \textbf{0.829} & \textbf{0.740} & \textbf{0.642} & 0.934 \\
\bottomrule
\end{tabular}
}
\caption{Molecule generation results on the PubChem324k dataset.}
\label{tab:pubchem_generation}
\end{table*}

\clearpage
\begin{table*}[ht]
\centering
\resizebox{0.7\linewidth}{!}{%
\begin{tabular}{@{}c|c@{}}
\toprule
Task & Template \\ \midrule
SMILES-to-caption & Provide a whole description of this molecule: <input> \\
Caption-to-SMILES & Provide a molecule based on this description: <input> \\
Substructure-to-phrase & Provide a keyword of this substructure: <input> \\
phrase-to-substructure & Provide a substructure based on this keyword: <input> \\ \bottomrule
\end{tabular}%
}
\caption{Prompt templates that are used for our multi-task pre-training of MolBridge-Gen.}
\label{tab:prompts}
\end{table*}

\begin{figure*}[ht]
    \centering
    \includegraphics[width=0.7\linewidth]{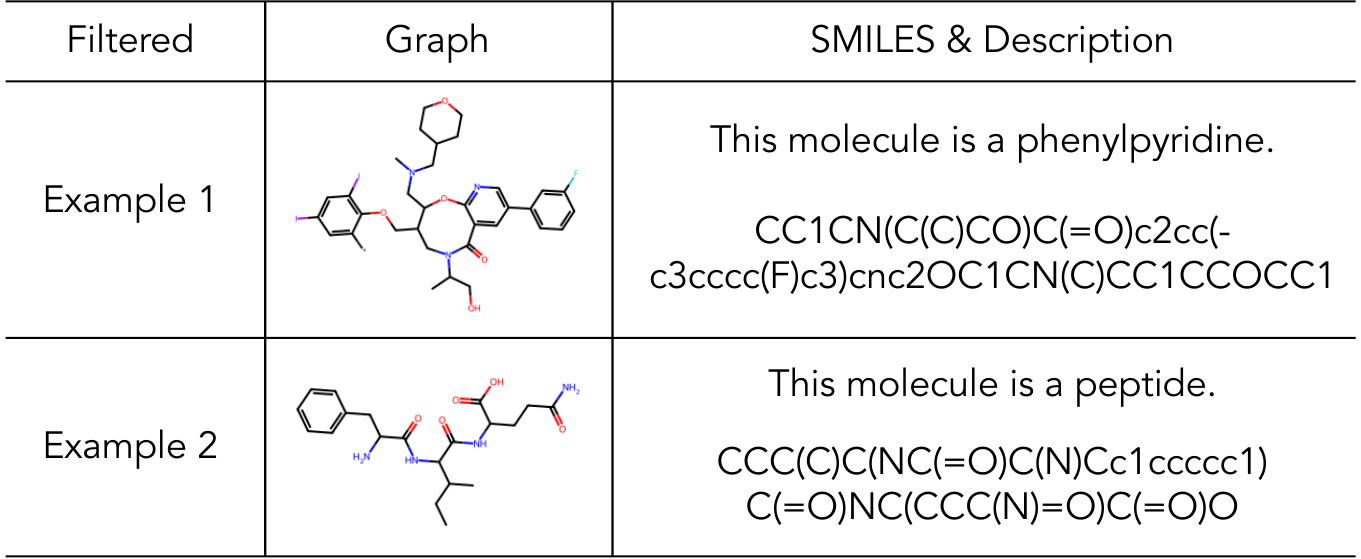}
    \caption{Examples of filtered augmented pairs.}
    \label{fig:filtered}
\end{figure*}

\begin{figure*}[ht]
    \centering
    \includegraphics[width=\linewidth]{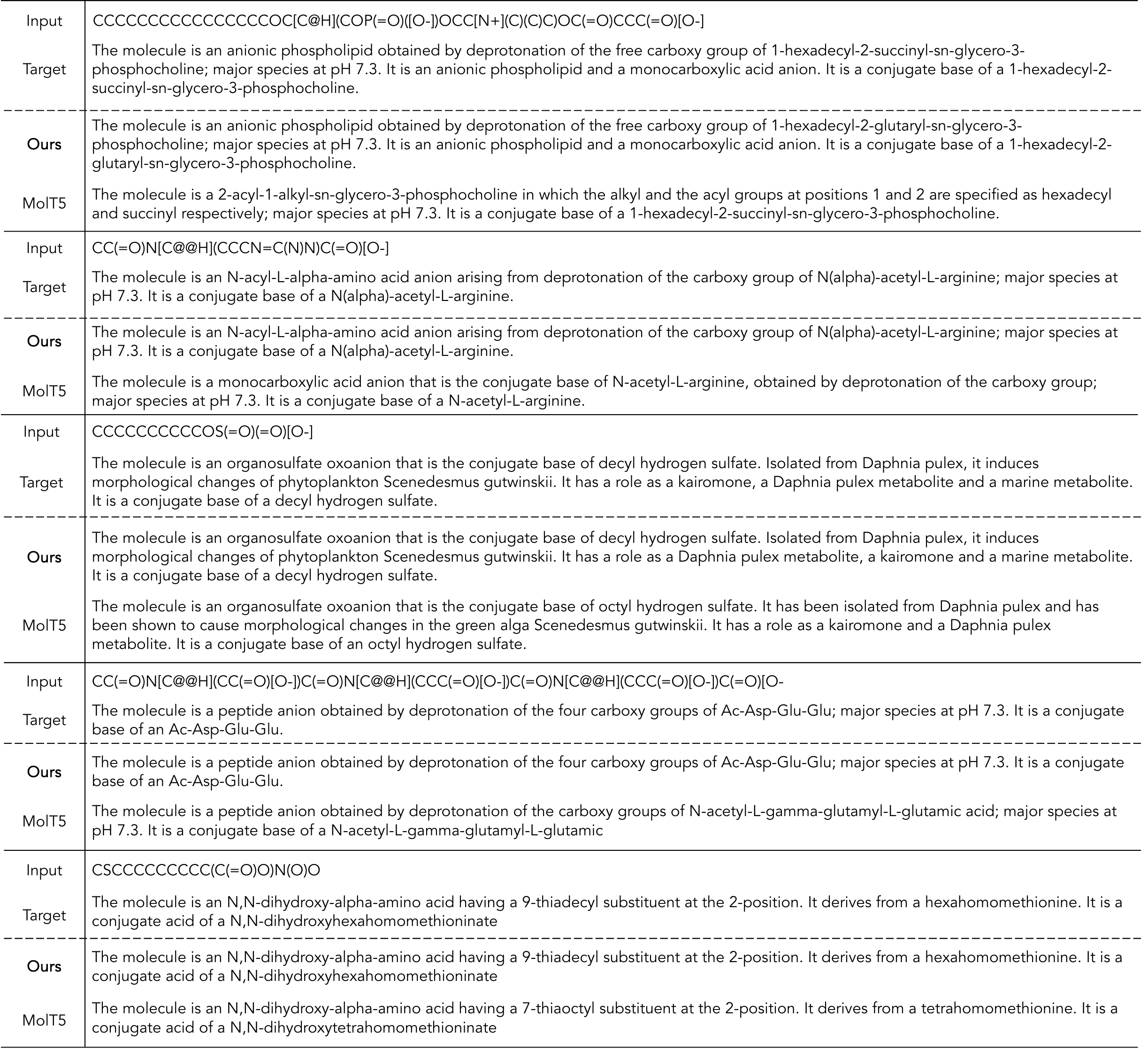}
    \caption{Examples of generated captions from input SMILES.}
    \label{fig:cap_ex}
\end{figure*}

\begin{figure*}[ht]
    \centering
    \includegraphics[width=\linewidth]{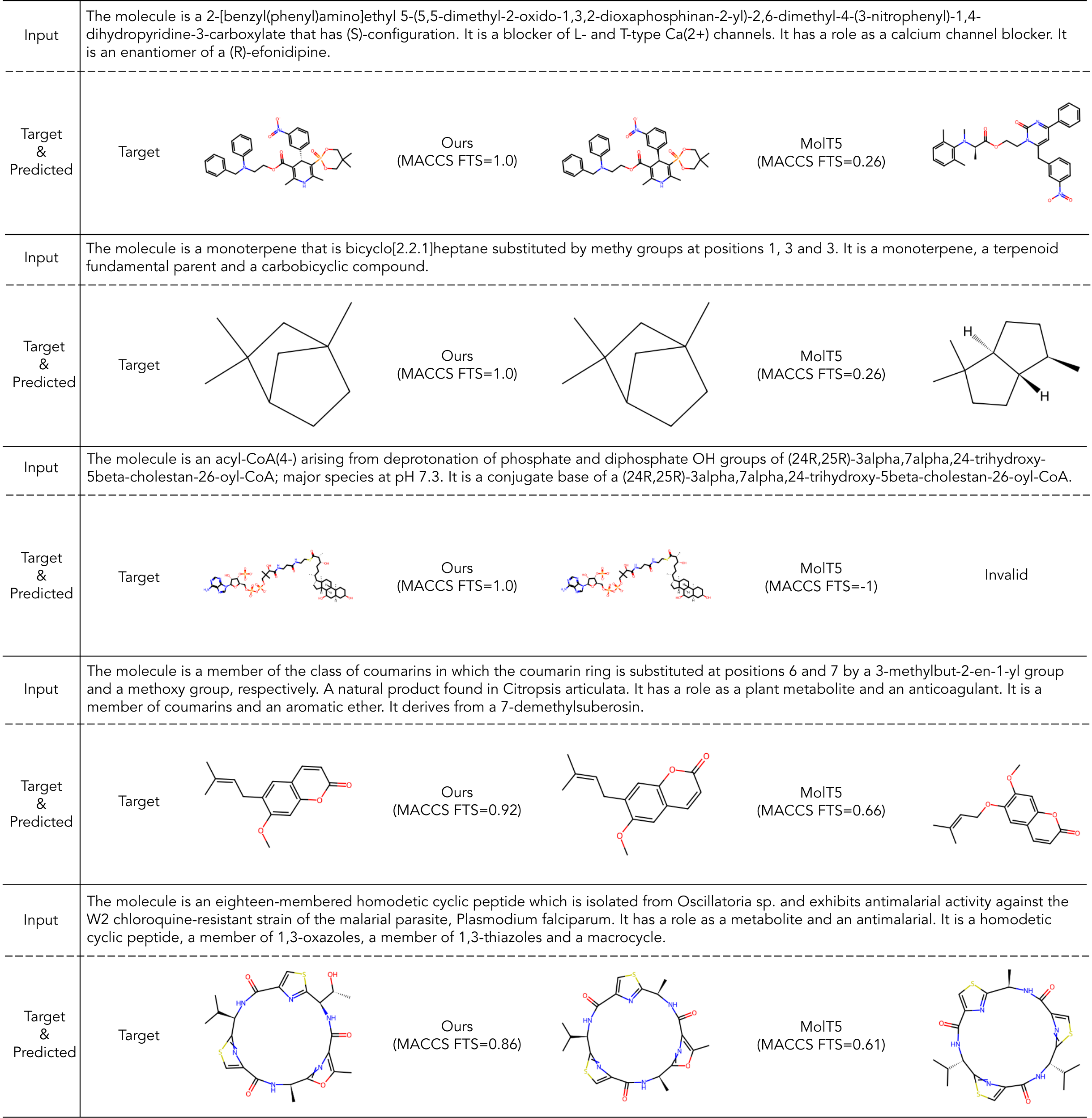}
    \caption{Visualized examples of generated SMILES from input caption.}
    \label{fig:mol_ex}
\end{figure*}

\begin{table*}[ht]
\centering
\resizebox{\textwidth}{!}{%
\begin{tabular}{@{}lccccccc@{}}
\toprule
\multicolumn{1}{l|}{Method} & \multicolumn{1}{c|}{\# Params} & BLEU-2$\uparrow$ & BLUE-4$\uparrow$ & ROUGE-1$\uparrow$ & ROUGE-2$\uparrow$ & ROUGE-L$\uparrow$ & METEOR$\uparrow$ \\ \midrule
\rowcolor{gray}
\multicolumn{8}{l}{1D SELFIES + 2D Graph} \\
\multicolumn{1}{l|}{Mol-Instructions \citep{DBLP:conf/iclr/FangL0LH0FC24-molinstruction}} & \multicolumn{1}{c|}{7B} & 0.249 & 0.171 & 0.311 & 0.203 & 0.239 & 0.271 \\
\multicolumn{1}{l|}{InstructMol-GS \citep{DBLP:conf/coling/CaoLLYL25-instructmol}} & \multicolumn{1}{c|}{7B} & 0.475 & 0.371 & 0.566 & 0.394 & 0.502 & 0.509 \\ \midrule
\rowcolor{gray}
\multicolumn{8}{l}{1D SELFIES + IUPAC name} \\
\multicolumn{1}{l|}{BioT5+ \citep{DBLP:conf/acl/PeiWGLFZ00024-biot5plus}} & \multicolumn{1}{c|}{252M} & 0.666 & 0.591 & \uline{0.710} & \uline{0.584} & \uline{0.650} & \uline{0.681} \\ \midrule
\rowcolor{gray}
\multicolumn{8}{l}{1D SMILES + 2D Graph + 2D Image} \\
\multicolumn{1}{l|}{GIT-Mol \citep{DBLP:journals/cbm/LiuRTR24}} & \multicolumn{1}{c|}{700M} & 0.352 & 0.263 & 0.575 & 0.485 & 0.560 & 0.533 \\ \midrule
\rowcolor{gray}
\multicolumn{8}{l}{1D SMILES + 2D Graph + Knowledge Graph} \\
\multicolumn{1}{l|}{MolFM-small \citep{molfm}} & \multicolumn{1}{c|}{136M} & 0.542 & 0.452 & 0.623 & 0.469 & 0.562 & 0.564 \\ 
\multicolumn{1}{l|}{MolFM-base \citep{molfm}} & \multicolumn{1}{c|}{296M} & 0.585 & 0.498 & 0.653 & 0.508 & 0.594 & 0.607 \\ \midrule
\rowcolor{gray}
\multicolumn{8}{l}{1D SMILES + 2D Graph} \\
\multicolumn{1}{l|}{MoMu-small \citep{momu}} & \multicolumn{1}{c|}{82M} & 0.532 & 0.445 & - & - & 0.564 & 0.557 \\ 
\multicolumn{1}{l|}{MoMu-base \citep{momu}} & \multicolumn{1}{c|}{252M} & 0.549 & 0.462 & - & - & 0.575 & 0.576 \\ 
\multicolumn{1}{l|}{MoMu-large \citep{momu}} & \multicolumn{1}{c|}{780M} & 0.599 & 0.462 & - & - & 0.593 & 0.597 \\ 
\multicolumn{1}{l|}{MolCA (Galactica-125M) \citep{molca}} & \multicolumn{1}{c|}{125M} & 0.612 & 0.526 & 0.674 & 0.521 & 0.606 & 0.636 \\ 
\multicolumn{1}{l|}{MolCA (Galactica-1.3B) \citep{molca}} & \multicolumn{1}{c|}{1.3B} & 0.620 & 0.531 & 0.681 & 0.537 & 0.618 & 0.651 \\ 
\multicolumn{1}{l|}{ICMA (Galatica-125M) \citep{DBLP:journals/corr/abs-2403-04197_icma}} & \multicolumn{1}{c|}{125M} & 0.636 & 0.565 & 0.674 & 0.536 & 0.615 & 0.648 \\
\multicolumn{1}{l|}{ICMA (Mistral-7B) \citep{DBLP:journals/corr/abs-2403-04197_icma}} & \multicolumn{1}{c|}{7B} & 0.651 & 0.581 & 0.686 & 
0.550 & 0.625 & 0.661 \\ \midrule
\rowcolor{gray}
\multicolumn{8}{l}{1D SMILES + Context Examples} \\
\multicolumn{1}{l|}{MolReFlect \citep{DBLP:journals/corr/abs-2411-14721}} & \multicolumn{1}{c|}{7B} & \textbf{0.676} & \textbf{0.608} & 0.703 & 0.571 & 0.644 & 0.680 \\ \midrule
\rowcolor{gray}
\multicolumn{8}{l}{1D SMILES} \\
\multicolumn{1}{l|}{MolT5-small \citep{molt5}} & \multicolumn{1}{c|}{77M} & 0.532 & 0.445 & 0.627 & 0.477 & 0.583 & 0.543 \\
\multicolumn{1}{l|}{MolT5-base \citep{molt5}} & \multicolumn{1}{c|}{248M} & 0.540 & 0.457 & 0.634 & 0.485 & 0.578 & 0.569 \\
\multicolumn{1}{l|}{MolT5-large \citep{molt5}} & \multicolumn{1}{c|}{783M} & 0.594 & 0.508 & 0.654 & 0.510 & 0.594 & 0.614 \\
\multicolumn{1}{l|}{Text+Chem T5 \citep{DBLP:conf/icml/Christofidellis23_t5+chem}} & \multicolumn{1}{c|}{220M} & 0.625 & 0.542 & 0.682 & 0.543 & 0.622 & 0.648 \\
\multicolumn{1}{l|}{MolXPT \citep{DBLP:conf/acl/LiuZXW0QZL23}} & \multicolumn{1}{c|}{350M} & 0.594 & 0.505 & 0.660 & 0.511 & 0.597 & 0.626 \\
\multicolumn{1}{l|}{MolReGPT (GPT-3.5) \citep{molregpt}} & \multicolumn{1}{c|}{-} & 0.565 & 0.482 & 0.623 & 0.450 & 0.543 & 0.585 \\
\multicolumn{1}{l|}{MolReGPT (GPT-4) \citep{molregpt}} & \multicolumn{1}{c|}{-} & 0.607 & 0.525 & 0.634 & 0.476 & 0.562 & 0.610 \\
\multicolumn{1}{l|}{Atomas-base \citep{zhang2025atomas}} & \multicolumn{1}{c|}{271M} & 0.632 & 0.545 & 0.685 & 0.545 & 0.626 & - \\
\multicolumn{1}{l|}{MolReFlect w/o Examples \citep{DBLP:journals/corr/abs-2411-14721}} & \multicolumn{1}{c|}{7B} & 0.617 & 0.539 & 0.657 & 0.510 & 0.593 & 0.623 \\
\rowcolor{lightblue}
\multicolumn{1}{l|}{MolBridge-Gen-small} & \multicolumn{1}{c|}{82M} & 0.625 & 0.542 & 0.686 & 0.549 & 0.629 & 0.649 \\ 
\rowcolor{lightblue}
\multicolumn{1}{l|}{MolBridge-Gen-base} & \multicolumn{1}{c|}{248M} & \uline{0.674} & \uline{0.605} & \textbf{0.724} & \textbf{0.609} & \textbf{0.676} & \textbf{0.693} \\ 
\bottomrule
\end{tabular}%
}
\caption{Results of molecule captioning task on CheBI-20 test set.}
\label{tab:molcap_appendix}
\end{table*}

\begin{table*}[ht]
\centering
\small
\resizebox{\textwidth}{!}{%
\begin{tabular}{@{}lccccccc@{}}
\toprule
Method & BLEU$\uparrow$ & EM$\uparrow$ & Levenshtein$\downarrow$ & MACCS FTS$\uparrow$ & RDK FTS$\uparrow$ & Morgan FTS$\uparrow$ & Validity$\uparrow$ \\ \midrule
\rowcolor{gray}
\multicolumn{8}{l}{1D SELFIES + IUPAC name} \\
BioT5+ \citep{DBLP:conf/acl/PeiWGLFZ00024-biot5plus} & 0.872 & \textbf{0.522} & 12.77 & 0.907 & 0.835 & 0.779 & \textbf{1.000} \\ \midrule
\rowcolor{gray}
\multicolumn{8}{l}{1D SMILES + 2D Graph + 2D Image} \\
GIT-Mol \citep{DBLP:journals/cbm/LiuRTR24} & 0.756 & 0.051 & 26.32 & 0.738 & 0.582 & 0.519 & 0.928 \\ \midrule
\rowcolor{gray}
\multicolumn{8}{l}{1D SMILES + 2D Graph + Knowledge Graph} \\
MolFM-small \citep{molfm} & 0.803 & 0.169 & 20.86 & 0.834 & 0.721 & 0.662 & 0.859 \\
MolFM-base \citep{molfm} & 0.822 & 0.210 & 19.45 & 0.854 & 0.758 & 0.758 & 0.892 \\ \midrule
\rowcolor{gray}
\multicolumn{8}{l}{1D SMILES + 2D Graph} \\
MoMu-small \citep{momu} & 0.800 & 0.150 & 21.45 & 0.818 & 0.709 & 0.651 & 0.858 \\
MoMu-base \citep{momu} & 0.815  & 0.183 & 20.52  & 0.847  & 0.737  & 0.678 & 0.863 \\
ICMA (Galatica-125M) \citep{DBLP:journals/corr/abs-2403-04197_icma} & 0.836 & - & 21.48 & 0.893 & 0.809 & 0.743 & 0.825 \\
ICMA (Mistral-7B) \citep{DBLP:journals/corr/abs-2403-04197_icma} & 0.855 & - & 18.73 & 0.916 & 0.837 & 0.789 & 0.891 \\ \midrule
\rowcolor{gray}
\multicolumn{8}{l}{1D SMILES + Context Examples} \\
MolReFlect \citep{DBLP:journals/corr/abs-2411-14721} & \textbf{0.903} 
 & \uline{0.510}  & \textbf{11.84}  & \textbf{0.929}  & \textbf{0.860}  & \textbf{0.813}  & 0.977 \\  \midrule
\rowcolor{gray}
\multicolumn{8}{l}{1D SMILES} \\
MolT5-small \citep{molt5} &  &  &  &  &  &  &  \\
MolT5-base \citep{molt5} & 0.779 & 0.082 & 25.19 & 0.788 & 0.662 & 0.602 & 0.787 \\
MolT5-large \citep{molt5} & 0.854 & 0.318 & 16.32 & 0.889 & 0.813 & 0.750 & 0.958 \\
Text+Chem T5 \citep{DBLP:conf/icml/Christofidellis23_t5+chem} & 0.853 & 0.322 & 16.87 & 0.901 & 0.816 & 0.757 & 0.943 \\
MolXPT \citep{DBLP:conf/acl/LiuZXW0QZL23} & - & 0.215 & - & 0.859 & 0.757 & 0.667 & 0.983 \\
MolReGPT (GPT-3.5) \citep{molregpt} & 0.790  & 0.139  & 24.91  & 0.847  & 0.708 & 0.624 & 0.887\\
MolReGPT (GPT-4) \citep{molregpt} & 0.857 & 0.280 & 17.14 & 0.903 & 0.805 & 0.739 & 0.899 \\
Atomas-base \citep{zhang2025atomas} & 0.868 & 0.343 & 13.76 & 0.908 & 0.827 & 0.773 & 0.971 \\
Atomas-large \citep{zhang2025atomas} & 0.874 & 0.387 & \uline{12.70} & 0.914 & 0.841 & 0.788 & 0.980 \\
MolReFlect w/o Examples \citep{DBLP:journals/corr/abs-2411-14721} & \uline{0.886} 
 & 0.430  & 13.99  & 0.916  & 0.828  & 0.775  & \uline{0.981} \\
\rowcolor{lightblue}
MolBridge-Gen-small & 0.827 & 0.266 & 16.88 & 0.898 & 0.820 & 0.751 & 0.947 \\ 
\rowcolor{lightblue}
MolBridge-Gen-base & 0.842 & 0.358 & 15.66 & \uline{0.918} & \uline{0.854} & \uline{0.798} & 0.956 \\
\bottomrule
\end{tabular}%
}
\caption{Results of text-based de novo molecule generation on CheBI-20 test set.}
\label{tab:molgen_appendix}
\end{table*}

\end{document}